\documentclass[10pt, a4paper]{article}
\usepackage{lrec}
\usepackage{multibib}
\usepackage{graphicx}
\usepackage{tabularx}
\usepackage{soul}
\usepackage{enumitem}
\usepackage{booktabs}
\usepackage{multirow}
\usepackage{enumitem}
\usepackage[ruled]{algorithm2e}
\SetAlFnt{\footnotesize}

\usepackage{epstopdf}
\usepackage[utf8]{inputenc}

\usepackage{hyperref}
\usepackage{xstring}

\title{Biomedical term normalization of EHRs with UMLS}

\name{Naiara Perez-Miguel, Montse Cuadros, German Rigau}

\address{HSLT group at Vicomtech, IXA group at UPV/EHU \\
         Donostia/San Sebasti\'an, Spain \\
         \{nperez,mcuadros\}@vicomtech.org, german.rigau@ehu.es}

\abstract{
This paper presents a novel prototype for biomedical term normalization of electronic health record excerpts with the Unified Medical Language System (UMLS) Metathesaurus, a large, multi-lingual compendium of biomedical and health-related terminologies. Despite the prototype being multilingual and cross-lingual by design, we first focus on processing clinical text in Spanish because there is no existing tool for this language and for this specific purpose. The tool is based on Apache Lucene\texttrademark\ to index the Metathesaurus and generate mapping candidates from input text. It uses the IXA pipeline for basic language processing and resolves lexical ambiguities with the UKB toolkit. It has been evaluated by measuring its agreement with MetaMap \textendash a mature software to discover UMLS concepts in English texts\textendash\ in two English-Spanish parallel corpora. In addition, we present a web-based interface for the tool.
\newline  }

\begin{document}

\maketitleabstract

\section{Introduction}
\label{sec:intro}

Biomedical text mining technologies are becoming a key tool for the efficient exploitation of information contained in unstructured data repositories, including scientific literature, Electronic Health Records (EHRs), patents, biobank metadata, clinical trials and social media. Natural Language Processing (NLP) and specifically Information Extraction (IE) tools, such as term normalization tools, can facilitate knowledge discovery, exchange, and reuse by finding relevant terms and semantic structure in those texts. This paper presents a preliminary application that enriches EHRs with links to the Unified Medical Language System (UMLS)\footnote{\href{https://www.nlm.nih.gov/research/umls/}{https://www.nlm.nih.gov/research/umls/}}, a multilingual repository of biomedical terminologies. The tool is multilingual and cross-lingual by design, but we first focus on Spanish EHR processing because there is no existing tool for this language and for this specific purpose. We propose a sequential pipeline that retrieves mapping candidates from an indexed UMLS Metathesaurus, uses the IXA pipeline \cite{Agerri2014} for basic language processing and UKB \cite{Agirre2009} for word sense disambiguation (WSD). In addition to the pipeline itself, this paper also presents a demonstration interface for the tool that will be available on-line\footnote{\href{http://demos-v2.vicomtech.org/umlsmapper/}{http://demos-v2.vicomtech.org/umlsmapper/}, user:vicomtech, password:umlsmapper}.

\section{Related Work}
\label{sec:related}

Biomedical term normalization is a long-established research field in English-speaking countries where terminological resources and basic-processing tools for the biomedical domain and this language have been available for decades. Thus, there already exist several mature applications that are being effectively exploited for different purposes and by different organizations as of today. In what follows, we present some of the better-known applications.

\textbf{MetaMap} \cite{Aronson2001,Aronson2006} enriches biomedical text with links to the UMLS Metathesaurus. It is ``knowledge intensive'' as it relies heavily on the SPECIALIST Lexicon, a large syntactic lexicon of biomedical and general English. \newcite{Meystre2005} evaluated MetaMap with 160 clinical documents of diverse nature (radiology reports, exam reports, and so on). MetaMap's results were compared to annotations by 8 physicians; the reported precision and recall for detecting a set of 80 diseases were 76\% and 74\%. 

\textbf{MedLEE} \cite{Friedman1994,Friedman2000} is one of the earliest English term mapping systems for the clinical domain, alongside MetaMap. It exploits several knowledge sources of their own. In \newcite{Friedman1994}, MedLEE is evaluated by measuring its precision and recall at detecting the presence of four diseases in a collection of health records; the results were 70\% recall and 87\% precision. 

\textbf{NCBO Annotator} is a web service provided by the National Center for Biomedical Ontology (NCBO) that annotates textual data with terms from the UMLS and BioPortal ontologies. The details of how MGREP ---the concept recognition tool--- works are limited to the conference poster by \newcite{Dai2008}. \newcite{Shah2009} experimented with the task of large-scale indexing of online biomedical resources: MetaMap recognized more concepts but with a lower precision than MGREP, and MGREP turned to be faster than MetaMap. 

\textbf{cTakes} \cite{Savova2010} is a comprehensive platform for performing many clinical information extraction tasks, including enriching text with terms from the UMLS Metathesaurus. cTakes does dictionary lookup to recognize and identify clinical entities. They report that mapping to the UMLS accuracy is high for exact span matches.

As for Spanish, there have been a few attempts to process clinical free text in this language. Next, we present some of these attempts that are relevant to the work presented in this paper. 

\textbf{GALEN} \cite{Carrero2008,Carrero2008a} proposed a ``Spanish MetaMap'' that combines machine translation techniques with the use of MetaMap. Unfortunately, they did not apply this system to any task, so performance scores cannot be reported. 

The system by \textbf{\newcite{Castro2010}} aims at retrieving SNOMED CT\textsuperscript{\textregistered} concepts based on an input phrase (SNOMED CT\textsuperscript{\textregistered} is the most complete biomedical terminology, and it is included in the UMLS). Term normalization is done by querying an Apache Lucene\texttrademark\ index of SNOMED CT\textsuperscript{\textregistered} and re-ranking the candidates with a function of their own. In order to evaluate the performance of this system, they obtained a set of 100 health records manually tagged by two specialists with ``disruptions'' or ``procedures'' concepts in SNOMED CT\textsuperscript{\textregistered}. For the exact-matching assessment, they report an average precision of 39\% and a recall of 0.65\%. Partial matching increases precision to 71\%, but recall is still 0.75\%. 

\textbf{FreelingMed} \cite{Oronoz2013} uses the Freeling analyzer \cite{Carreras2004} and extend its linguistic data with various knowledge sources including SNOMED CT\textsuperscript{\textregistered}, a list of medical abbreviations \cite{Yetano2003}, Bot PLUS, and ICD-9. The actual task that the tool is meant to perform is term recognition, not term normalization. The system was assessed against a Gold Standard of 100 health records annotated with drug names, diseases and substances, counting as true positives approximate matches. The final result was 0.90 per the F-measure.

As can be seen, none of the tools presented offers a complete pipeline to perform biomedical term normalization in Spanish clinical text with the UMLS.

\begin{figure*}[h]
  \centering
    \includegraphics[width=0.8\textwidth]{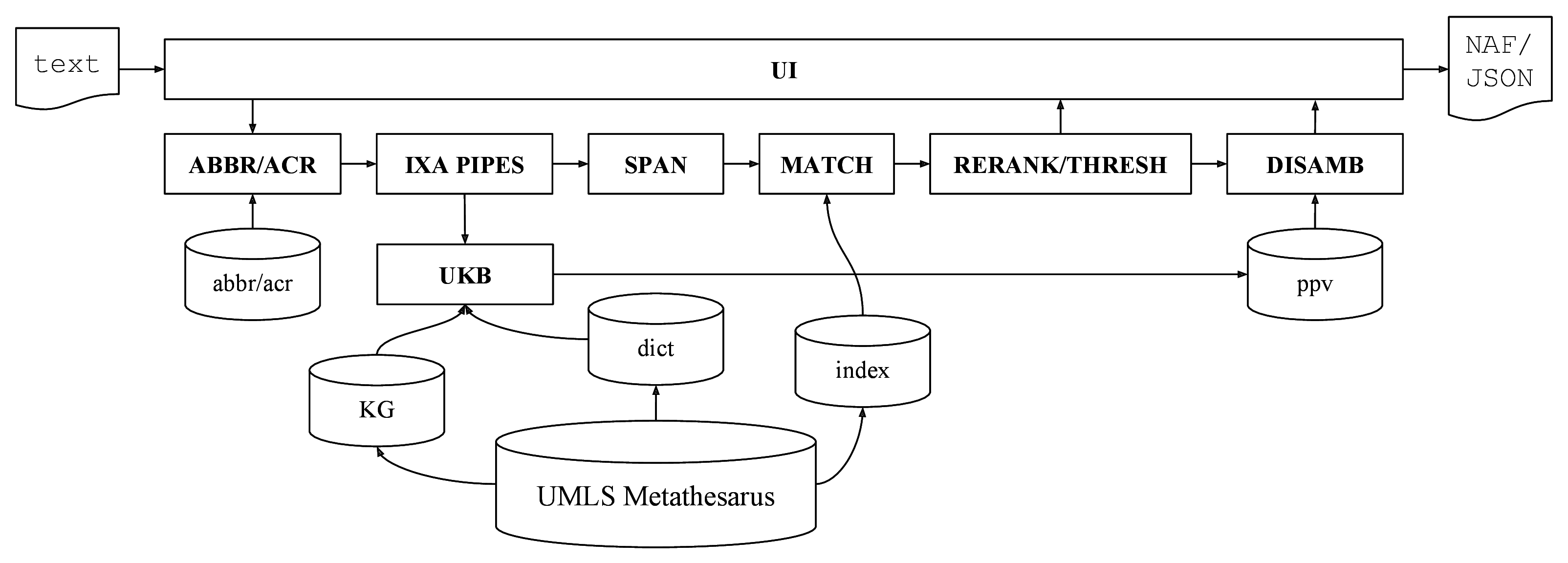}
    \caption{Architecture of the pipeline}
    \label{fig:architecture}
\end{figure*}

\section{Pipeline Description}
\label{sec:our}

The overall architecture for the prototype is schematized in Figure \ref{fig:architecture}. It consists of components executed in sequence, some of which use a knowledge base, our adaptation of the UMLS Metathesaurus. This section provides a description of the knowledge base and the overall workflow. We also report a first approximation for assessing the performance of the prototype.

\subsection{The Knowledge Base}
\label{ssec:kb}

The knowledge base of the prototype has been derived from the 2016AA Full Release UMLS Metathesaurus. It gathers 196 terminology sources in 25 different languages, amounting to 3,250,226 concepts and 10,586,865 unique terms in total. For this prototype we focus on the subset of sources in Spanish, which consists of 451,297 concepts and 1,255,377 unique terms. Table \ref{tab:countspersource} shows the amount of concepts and unique terms per source available in Spanish \textendash 7 out of 196 \textendash, both in their English and Spanish versions. The table reveals that the Spanish versions have much less conceptual and lexical coverage.  

To build the knowledge base for our prototype, we use specifically Metathesaurus terms that \textit{a)} are in Spanish, \textit{b)} do not belong to LOINC\textsuperscript\textregistered\ \footnote{LOINC\textsuperscript\textregistered\ descriptors look typically like ``especie de Thrichomonas:n\'umero areico:punto en el tiempo:sedimento urinario:cuantitativo:microscopia.de luz.campo de gran aumento'', so they are not suited for the task at hand.}, \textit{c)} are shorter than 15 tokens, \textit{d)} are not obsolete or suppressible, \textit{e)} do not consist of a single character, \textit{f)} do not consist of just numbers, and \textit{g)} do not consist of only stopwords. We consider 303 common Spanish words except ``no'', ``sin'' and ``con'' (\textit{no}, \textit{without}, and \textit{with}, respectively) because they may alter the polarity of expressions, which is essential to be processed in this domain \cite{Ceusters2007}. Applying these filters, we are left with 352,075 concepts and 546,309 unique terms. 

The application proposed needs the knowledge base in three formats:

\begin{table}
	\centering
    \small
	\begin{tabular}{@{}lrrrr@{}}
		\toprule
		\textbf{}                                       & \multicolumn{2}{c}{\textbf{English}}                                       & \multicolumn{2}{c}{\textbf{Spanish}}                                       \\ \cmidrule(l){2-5} 
		\textbf{}                                       & \multicolumn{1}{c}{\textbf{Concepts}} & \multicolumn{1}{c}{\textbf{Terms}} & \multicolumn{1}{c}{\textbf{Concepts}} & \multicolumn{1}{c}{\textbf{Terms}} \\ \midrule
		\textbf{All sources}                            & 3,250,226                             & 10,586,865                           & 451,297                               & 1,255,377                           \\
		\textbf{CPT\textsuperscript{\textregistered}}   & 39,152                                & 61,923                              & 2,720                                 & 2,484                               \\
		\textbf{ICPC}                                   & 748                                   & 1,017                               & 722                                   & 688                                 \\
		\textbf{LOINC\textsuperscript{\textregistered}} & 157,645                               & 390,425                             & 48,609                                & 48,631                              \\
		\textbf{MedDRA}                                 & 51,961                                & 78,528                              & 45,488                                & 61,103                              \\
		\textbf{MeSH\textsuperscript{\textregistered}}  & 359,116                               & 837,305                             & 35,970                                & 64,804                              \\
		\textbf{SCT\textsuperscript{\textregistered}}   & 357,448                               & 1,115,865                           & 306,539                               & 746,600                             \\
		\textbf{WHO-ART}                                & 3,175                                 & 3,831                               & 2,566                                 & 3,102                               \\ \bottomrule
	\end{tabular}
	\caption{UMLS 2016AA Full Release Metathesaurus counts for English and Spanish subsets of sources available in Spanish}
	\label{tab:countspersource}
\end{table}

\textbf{The UMLS index.} We use Apache Lucene\texttrademark\ in order to be able to make fast searches in our subset of the UMLS Metathesaurus. An index has been created where each entry represents a term of the subset and contains the following information: the term itself, a normalized version of the term, the concept identifier(s) it is related to, and its source(s). The normalized string is obtained after erasing spurious parenthetical content, punctuation, and stopwords. The list of the spurious parenthetical content has been curated manually after studying the Metathesaurus. As for the stopwords, they are the same 303 used to filter the UMLS Spanish subset.

\textbf{The UKB Knowledge Graph.} This graph contains all the relations in the 2016AA Metathesaurus whose origin and target concepts are both included in our UMLS subset. For each relation, it encodes the source and target concepts, the direction of the relation, and its type. Overall, the graph consists of 352,075 vertices and 8,381,482 edges. All the concepts indexed participate in one relation at least.

\textbf{The UKB Dictionary.} It maps the terms in our UMLS subset to their respective concept or concepts, in the case of those that are ambiguous.

\subsection{Overview of the Workflow}
\label{ssec:wf}

Let us describe the proposed processing flow by means of an example; take the input text to be the following:
\begin{center}
	``acude por lesión grave en rodilla dcha''\\
    \textit{\small [patient] comes due to serious injury in rt knee}
\end{center} 
First, the text received is analyzed in search of \textbf{abbreviations and acronyms}, which are expanded to their corresponding full expressions. The tool employed to identify abbreviation- or acronym-like elements in texts \cite{Montoya2017} exploits a set of rules and a 2,312-item long list of abbreviation/acronym and corresponding expansions, curated after manual annotations by health care professionals. In our example, this step would produce
\begin{center}
	``acude por lesión grave en rodilla derecha''\\
    \textit{\small [patient] comes due to serious injury in right knee}
\end{center} 
Next, the system does basic \textbf{linguistic processing} with the IXA pipeline \cite{Agerri2014}: tokenization, part-of-speech tagging, and constituent parsing. The linguistic information obtained serves as basis to perform \textbf{boundary detection}, that is, to recognize in the text spans or sequences of tokens that are likely to be mapped to a medical concept. In order to maximize recall, 
we explore two methods: extracting n-grams of varying sizes, and extracting nominal phrases based on a simple set of rules that uses the linguistic information, allowing for discontinuous spans.

After extracting textual spans, the system attempts to find mapping candidates of the Metathesaurus terms indexed by lexical proximity. This is the role of the \textbf{matching} module. It queries the index with the spans, obtaining as a result of each query a collection of Metahesaurus terms, which are in turn related to one or more concepts and a relevance score. 

The \textbf{reranking} module assigns new scores to the candidates using a function other than the one provided by Lucene. We explore two such functions: the one by \newcite{Castro2010}, and the one by \newcite{Aronson2001} implemented in MetaMap. Furthermore, a \textbf{threshold} can be applied to discard candidates with low scores.

\begin{figure}[h]
  \centering
  \includegraphics[width=0.9\linewidth]{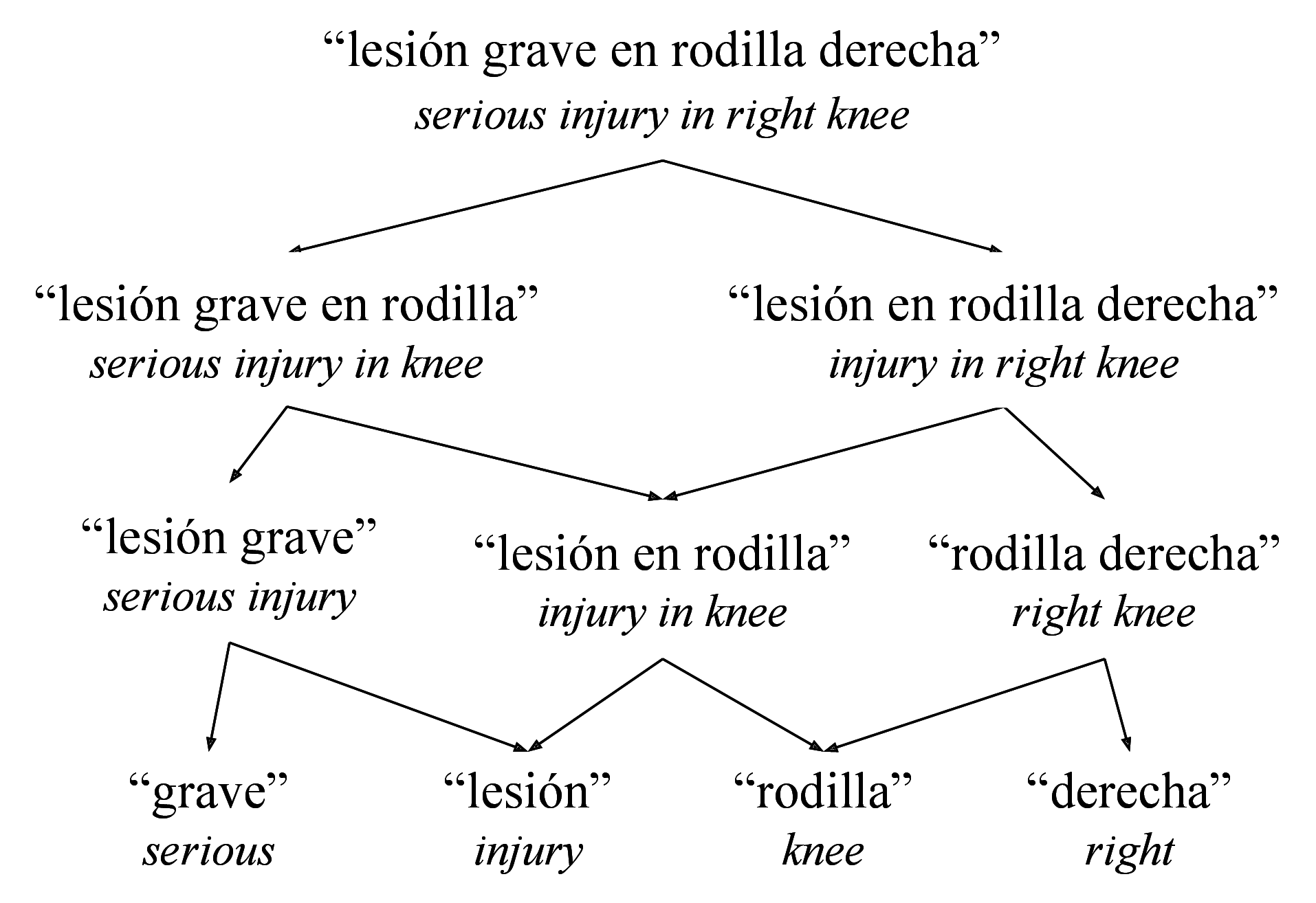}
  \caption{Oriented tree of detected spans}
  \label{fig:chunk-order}
\end{figure}

Matching, reranking and thresholding are not done with all the spans detected; the \textbf{mapping candidate generation algorithm} prefers longer matches: 
\begin{enumerate}
\item the system orders the spans by subsumption creating oriented trees as depicted in Figure \ref{fig:chunk-order};
\item then, it queries the index with the root of the tree and its direct children, reranks the results and applies a threshold;
\item if any of the children obtains a better result than their parent, then the results retrieved for the parent span are ruled out, and the algorithm is repeated recurrently for the children nodes;
\item if a parent has a result better than any of its children's, the results retrieved for the parent are accepted as candidates and the system does not attempt to map any of its descendants. 
\end{enumerate}
Following this algorithm, textual spans that overlap can be annotated with different concepts, but not spans that are nested within a bigger one.

At this point, a span can have zero, one or multiple mapping candidates. Then, 
\begin{enumerate}[label=\alph*)]
\item if no candidate is available, one must conclude that either the span in question was never a term in the first place, or that it is a term but does not have an explicit or convincing enough mapping to a UMLS Metathesaurus entry indexed;
\item if one candidate is available, the system takes it as a final mapping for the span;
\item if more than one is available, the system takes as a final mapping the one scored highest; and
\item if more than one candidate become tied in first position, the system needs to carry out a \textbf{disambiguation} step in order to choose the correct mapping. This process is performed by the UKB module.
\end{enumerate}

The algorithm behind UKB is \textit{Personalized PageRank} \cite{Haveliwala2002}. \newcite{Agirre2010} and \newcite{Stevenson2012} prove that UMLS's conceptual graph can be used as a knowledge base for WSD. Here we implement a little variation of their approach. The context to initialize the Knowledge Graph consists of the tokens in the text; the system is able to provide this information as early as the basic linguistic processing is done. When the disambiguation module is required, it just needs to choose the mapping candidate with highest activation in the Personalized PageRank Vector.

The pipeline ends by gathering the final mappings and displaying them to the user.

\subsection{Evaluation}
\label{ssec:eval}

At the moment there is no corpus available in Spanish annotated with UMLS concepts that can serve as Gold Standard to evaluate this application. For this reason, we propose the following evaluation framework as a first approximation to measure the performance of the tool proposed. 

\subsubsection{Design}
\label{sssec:design}

Having created/obtained two English-Spanish parallel corpora of biomedical text, the English documents have been annotated with MetaMap and the Spanish ones with the prototype proposed; then, the agreement between the systems has been measured by means of Cohen's Kappa \cite{Cohen1960}. Crucially, MetaMap'a knowledge source has been reduced so that both systems can annotate only the same 352.075 concepts, in order to make the annotations comparable. MetaMap's mapping strategy has also been configured so that it prefers longer matches, as the prototype does.

\paragraph{Corpora.} One of the corpora is a manually revised subset of the Scielo Corpus \cite{Neves2016}, resulting in 1,895 titles and abstracts of scientific literature. The other corpus consists of 10 EHR texts originally in Spanish and their English translations, plus 8 EHR texts originally in English and their translations to Spanish. Table \ref{tab:corp} shows the sizes of the corpora.

\begin{table}[h]
	\centering
	\begin{tabular}{@{}rcrcr@{}}
		\toprule
		\multicolumn{1}{l}{} & \multicolumn{2}{c}{\textbf{Scielo}} & \multicolumn{2}{c}{\textbf{EHR}} \\ \midrule
		& \textbf{es} & \multicolumn{1}{c}{\textbf{en}} & \textbf{es} & \multicolumn{1}{c}{\textbf{en}} \\ \midrule
		\# documents & \multicolumn{1}{r}{1,895} & 1,895 & \multicolumn{1}{r}{18} & 18 \\
		\# words & \multicolumn{1}{r}{26,490} & 23,374 & \multicolumn{1}{r}{23,311} & 21,093 \\ \bottomrule
	\end{tabular}
	\caption{Corpora used for evaluation}
	\label{tab:corp}
\end{table}

\paragraph{Metric.} Cohen's Kappa $k$ is defined as follows:

\begin{equation}
k = \frac{p_o - p_e}{1 -p_e}
\end{equation}

\noindent where $p_o$ is the proportion of units in which the annotators agree and $p_e$ is the proportion of units for which agreement is expected by chance. The units are the 352.075 concepts in the index; MetaMap and our system agree only when both say that a given concept is present in the input document.

There is no universally accepted interpretation of Cohen's kappa as to what is considered high or low agreement. \cite{landis1977} proposed the following, which is widely cited, but has no evidential grounding:

\begin{table}[h!]
  \centering
  \begin{tabular}{@{}rrrl@{}}
  $k < 0.00 $  & & & No agreement \\
  $0.00 \leq k \leq 0.20$ & & & Slight agreement \\
  $0.21 \leq k \leq 0.40$ & & & Fair agreement\\
  $0.41 \leq k \leq 0.60$ & & & Moderate agreement \\
  $0.61 \leq k \leq 0.80$ & & & Substantial agreement \\
  $0.81 \leq k \leq 1.00$ & & & Almost perfect agreement \\
  \end{tabular}
\end{table}

\paragraph{Variables.} The experiment has been carried out with the following  prototype settings:

\begin{itemize}[noitemsep]
\item Boundary detection: \textit{ngram} or \textit{phrase}.
\item Re-ranking function: Lucene (\textit{L}), \newcite{Castro2010} (\textit{C}), or \cite{Aronson2001} (\textit{A}); \textit{L} is simply using the scores given by Lucene, that is, not re-ranking at all.
\item Disambiguation: \textit{UKB} or random disambiguation as baseline (\textit{rand}).
\end{itemize}

The results reported can only be taken as hints for the differences in performance between the possible configurations of the modules. Therefore, a qualitative error and disagreement analysis has been carried out in an attempt to elucidate these issues.

\begin{table}[h]
	\centering
	\begin{tabular}{@{}ccccc@{}}
		\toprule
		& \textbf{WSD} & \textbf{score} & \textbf{ngram} & \textbf{phrase} \\ \midrule \midrule
	\multirow{6}{*}{Scielo}	& \multirow{3}{*}{rand}                                                  & L$_{(.0)}$                                                              & 0.323 $\pm$ 0.006                                                                      & 0.304 $\pm$ 0.006                                                                       \\
		& & A$_{(.5)}$                                                             & 0.331 $\pm$ 0.006                                                                      & 0.308 $\pm$ 0.006                                                                       \\
		& & C$_{(.7)}$                                                              & 0.398 $\pm$ 0.006                                                                      & 0.372 $\pm$ 0.006                                                                       \\ \cmidrule(l){2-5}
		& \multirow{3}{*}{UKB}                                                     & L$_{(.0)}$                                                              & 0.343 $\pm$ 0.006                                                                      & 0.328 $\pm$ 0.005                                                                       \\
		& & A$_{(.5)}$                                                             & 0.349 $\pm$ 0.006                                                                      & 0.330 $\pm$ 0.006                                                                       \\
		& & C$_{(.7)}$                                                              & \textbf{0.412 $\pm$ 0.006}                                                                      &                0.387 $\pm$ 0.006                                                            \\ \midrule \midrule
        \multirow{6}{*}{EHR} & \multirow{3}{*}{rand}                                                  & L$_{(.0)}$                                                              & 0.286 $\pm$ 0.007                                                                      & 0.266 $\pm$ 0.007                                                                       \\
		& & A$_{(.5)}$                                                             & 0.330 $\pm$ 0.008                                                                      & 0.316 $\pm$ 0.008                                                                       \\
		& & C$_{(.7)}$                                                              & 0.403 $\pm$ 0.008                                                                      & 0.389 $\pm$ 0.008                                                                       \\ \cmidrule(l){2-5}
		& \multirow{3}{*}{UKB}                                                     & L$_{(.0)}$                                                              & 0.321 $\pm$ 0.007                                                                      & 0.306 $\pm$ 0.007                                                                       \\
		& & A$_{(.5)}$                                                             & 0.365 $\pm$ 0.008                                                                      & 0.354 $\pm$ 0.008                                                                       \\
		& & C$_{(.7)}$                                                              & \textbf{0.432 $\pm$ 0.008}                                                                      &                0.414 $\pm$ 0.008                                                           \\ \bottomrule
	\end{tabular}
    \caption{Agreement between MetaMap and the prototype}
	\label{tab:results}
\end{table}

\subsubsection{Results}
\label{sssec:results}
Results show that our prototype can reach moderate agreement with MetaMap. They suggest that the scoring function proposed in \newcite{Castro2010} makes the results of our prototype substantially more similar to the ones from MetaMap than the other two functions. Using n-grams to create textual spans yields always a slightly better agreement with MetaMap. Furthermore, agreement with MetaMap also improves when using UKB to perform disambiguation compared to the baseline proposed. 

A manual analysis of the results has shown that the main source of \textbf{disagreement} is, of course, the fact that MetaMap and our application annotate different texts --parallel texts; furthermore, they use different sources of knowledge, in spite of the efforts to make them as similar as possible by limiting the knowledge base of MetaMap to contain only the concepts indexed for our system. To illustrate these facts, let us consider the following input:\\

\noindent en: ``Should we rule out congenital anesplenia?''\\
\noindent es: ``¿Debemos descartar una asplenia cong\'enita?''\\

\noindent MetaMap and our best system (\textit{UKB}+\textit{C}+\textit{ngram}) find mappings for these spans:\\

\noindent MetaMap: ``rule'', ``out'', ``congenital''\\
\noindent Ours: ``descartar'', ``asplenia cong\'enita''\\

To begin with, ``rule out'' is translated as ``descartar'' in Spanish. When MetaMap creates ---in this case, incorrect--- annotations for ``rule'' and ``out'', it is impossible to produce the same annotations, since the Spanish ``descartar'' does not have the meaning of any of the two English words separately. We can also see that MetaMap does not recognize the concept ``congenital anesplenia''. As it happens, MetaMap's knowledge base contains ``congenital asplenia'' but not ``congenital anesplenia'', and so it does not annotate it. Of course, problems like these occur in both directions.

As for the errors that our prototype commits, many \textbf{false positive errors} are produced due to the fact that the Metathesaurus does not capture all the possible meanings of the terms it contains; because candidates are scored simply by means of lexical similarity, the system will annotate a term that is similar enough to an entry in the Metathesaurus even if they denote different concepts. Let us illustrate the problem: the term ``clavo'' in Spanish has at least three meanings: \textit{a)} clove (a spice), \textit{b)} nail or rod (a metallic object), and \textit{c)} corn of toe (a disease). However, the term ``clavo'' is only related to sense \textit{a)} and \textit{c)} in the Metathesaurus. This is not to say that sense \textit{b)} is not represented in the Spanish subset, but that it is not represented as ``clavo''. As a consequence, whenever an input text contains ``clavo'' (and it does not form a bigger concept with its surrounding words), it will be annotated as being a disease or a spice, even if it is neither of the two.

Another important source of false positives is the over-generation of spans: both n-gram-based and phrase-based detection generate incorrect spans that eventually can also be annotated. The n-gram strategy clearly generates spans that are not meant to form syntactic units, and thus neither intended meaning units. For example, in the text fragment ``[...] arteria tor\'acica en radiograf\'ia [...]'' (\textit{chest artery in x-ray}), the bigram [tor\'acica, radiograf\'ia] would form a span that would, in turn, trigger mapping candidates consisting of concepts referring to chest x-ray, which is not actually mentioned in the text. Although the phrase-based strategy was meant to overcome this problem by leveraging syntactic information, the fact that it allows for discontinuous spans also produces over-generation sometimes, especially when coordination and/or enumeration are involved.

Regarding \textbf{false negatives}, there are two main reasons for our system to miss a biomedical concept: on the one hand, it can happen that the concept is not captured in the Metathesaurus at all; on the other hand, it could be that the concept is captured but not as expressed in the text, be it because it is misspelled, abbreviated in a way that the Metathesaurus does not contemplate, or formulated in any other non-standard way. That is, false negatives are caused by a poor lack of the Metathesaurus and the lexical variability of clinical narrative. MetaMap relies on a powerful tool to deal with variability \textendash the SPECIALIST Lexicon; we do not address variability but for a closed list of abbreviations. As a consequence, our system is much more likely to produce this type of error, in any of its possible configurations. 

Additionally, phrase-based span detection is another source of false negatives, as it can miss noun phrases due to errors in the lower-level processing of the input texts: if it misses a noun phrase and the noun phrase happens to be a relevant term, the term is not annotated.

\section{Demo}
\label{sec:demo}

\begin{figure*}[h]
  \centering
    \includegraphics[width=\textwidth]{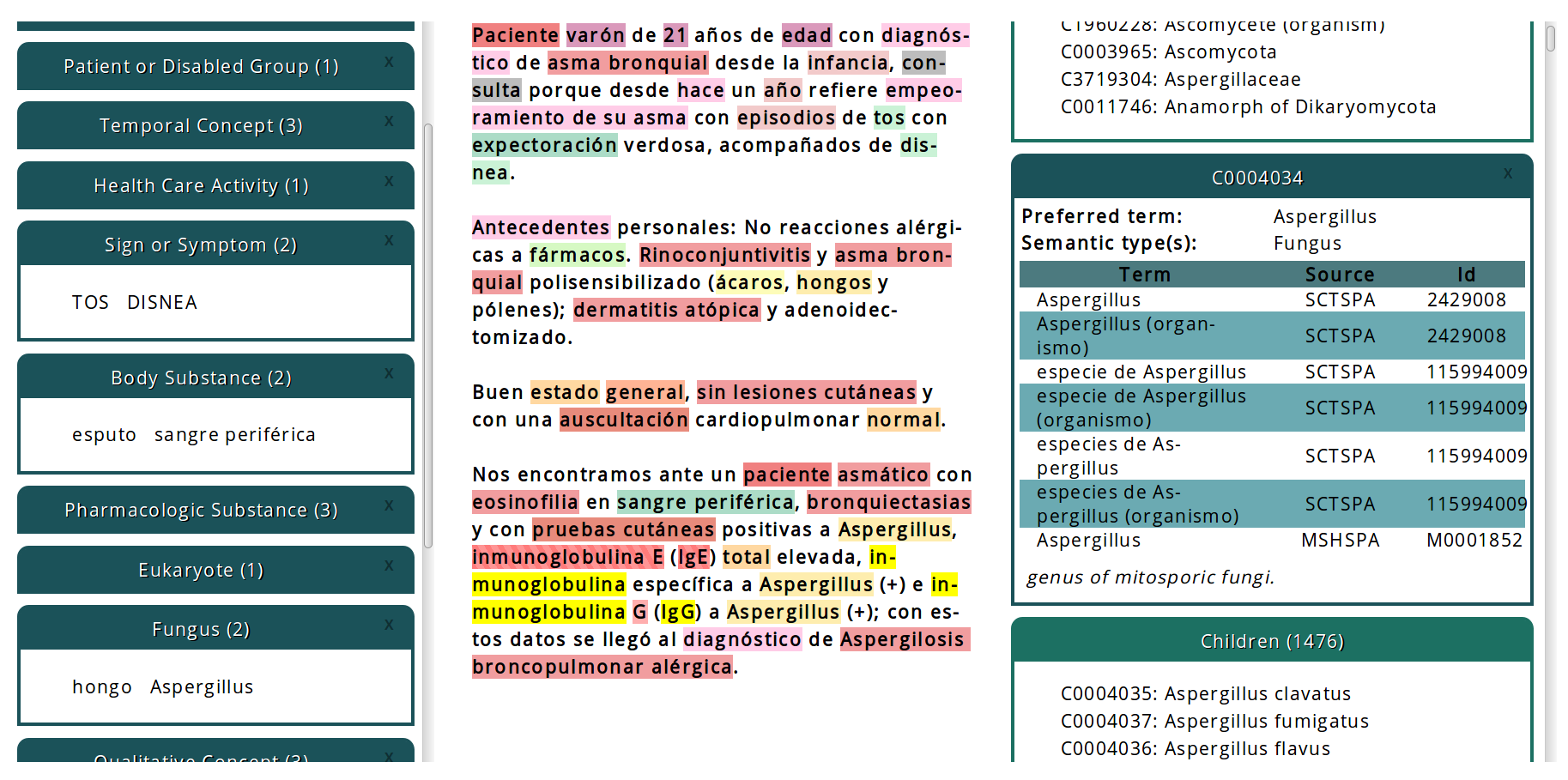}
    \caption{Results page of the demo website}
    \label{fig:demo}
\end{figure*}

A web-based demonstrator has been developed to allow users to introduce a text of their choosing and visualize the mappings produced by the application in an interactive user interface. The client side of the demonstrator has been developed in Angular2\footnote{\url{https://angular.io/}}. It is a webservice that communicates with the application via HTTP. In order to enrich the demonstrator with information about the concepts that have been mapped, the demonstrator also communicates with an additional webservice that provides an API to query the UMLS Metathesaurus and the Semantic Network, which is a hierarchical classification of the concepts in the UMLS Metathesaurus, and a source of the Metathesaurus itself.

In the \textbf{home page}, users can introduce their text and configure the application. Users can also choose which semantic types of the Semantic Network of the UMLS they are interested in; the bottom part of the page contains the whole Semantic Network in the form of a tree that can be expanded and collapsed by the users in order to select the semantic types of the concepts to be used by the mapping procedure. 

\textbf{The result page} is divided into three columns. An example is shown in Figure \ref{fig:demo}. The middle column contains the submitted text; annotations are marked in the text with different colors, depending on the semantic type of the concepts. On the left side is a list of the found concepts' semantic types. By clicking on any of the semantic types, one can see below the actual concepts or annotations, represented by their preferred names. The example given in Figure \ref{fig:demo} shows, for instance, that two signs or symptoms haven been found in the text (i.e. ``tos'' \textendash \textit{cough}\textendash\ and ``disnea'' \textendash \textit{dyspnea}\textendash ). When the user clicks on one of the concept names, information about that concept appears on the right side of the page: preferred name, semantic types, a definition, and so on. Moreover, the user can also see hypernym and hyponym relations, and navigate through the concepts within this hierarchy. In the case of Figure \ref{fig:demo}, the user clicked on the concept ``Asperguillus'' \textemdash which is mentioned twice in the last paragraph of the processed text\textemdash. The figure shows that this concept, with identifier C0004034 in the UMLS, has 6 terms related to it in the Spanish extension of SNOMED-CT (SCTSPA) and one more in the Spanish translation of Medical Subject Headings (MSHSPA).  It also shows, among other information, that ``Aspergillus'' is a ``Ascomycota'', and that ``Aspergillus clavatus'', ``Aspergillus fumigatus'' and ``Aspergillus flavus'' are all ``Aspergillus''.

\section{Conclusions}
We have presented a prototype to perform biomedical term normalization in clinical texts with the UMLS Metathesaurus. The tool performs abbreviation/acronym expansion and WSD. Mapping candidate generation is done by querying an index of the Metathesaurus with spans of the input text. As a preliminary evaluation, agreement with MetaMap has been measured in two parallel corpora; our best system has reached moderate agreement with MetaMap. We have also presented a web-based user interface for the prototype. As future work, we plan to assess the tool with texts in languages other than Spanish. We must also address misspellings, morphological variants and synonyms of the terms covered in the UMLS. Furthermore, other evaluation frameworks for evaluation should be designed, in order to better understand the shortages that the current version of the prototype has and how the tool could be improved.

\section{Acknowledgements}
This work has been funded by the Department of Economic Development and Infrastructure of the Basque Government under the project BERBAOLA (KK-2017/00043), and by the Spanish Ministry of Economy and Competitiveness (MINECO/FEDER, UE) under the projects CROSSTEXT (TIN2015-72646-EXP)
and TUNER (TIN2015-65308-C5-1-R).

\section{Bibliographical References}
\label{main:ref}

\bibliographystyle{lrec}
\bibliography{xample}


\end{document}